\documentclass[10pt, runningheads]{llncs}
\usepackage{listings}
\usepackage{amsmath,amssymb,amsfonts}
\usepackage{algorithm}
\usepackage{algpseudocode}
\usepackage{graphicx}
\usepackage{textcomp}
\usepackage{float}
\usepackage[dvipsnames,table]{xcolor}
\def\BibTeX{{\rm B\kern-.05em{\sc i\kern-.025em b}\kern-.08em
    T\kern-.1667em\lower.7ex\hbox{E}\kern-.125emX}}
    
\usepackage{subcaption}

\usepackage{hyperref}
\usepackage{capt-of}
\usepackage{listings}
\usepackage[style=ieee,maxnames=1,minnames=1,backend=biber,bibstyle=ieee,citestyle=numeric-comp]{biblatex}
\usepackage{url}
\usepackage{tcolorbox}
\usepackage{rotating}
\usepackage{textpos}
\usepackage{multirow}

\newcounter{obsno}
\definecolor{textHighlight}{RGB}{227,242,253}

\begin{document}
\renewcommand{\thelstlisting}{\arabic{lstlisting}}


\title{Adaptive PCA-Based Outlier Detection for Multi-Feature Time Series in Space Missions\thanks{This work is supported by the European Commission, with Automatics in Space Exploration (ASAP), project no. 101082633.}
}
\titlerunning{Adaptive PCA-Based Outlier Detection for Space Missions}
\authorrunning{J. Ekelund at al.}
\author{Jonah Ekelund$^1$, Savvas Raptis$^2$,
Vicki Toy-Edens$^2$,
Wenli Mo$^2$,
Drew L. Turner$^2$,
Ian J. Cohen$^2$,
Stefano Markidis$^1$}
\institute{
\textit{$^1$KTH Royal Institute of Technology, Computer Science Dept., Stockholm, Sweden}\\
\textit{$^2$Johns Hopkins University Applied Physics Laboratory, Laurel, MD, USA}\\
}

\maketitle
\begin{abstract}
    Analyzing multi-featured time series data is critical for space missions making efficient event detection, potentially onboard, essential for automatic analysis. However, limited onboard computational resources and data downlink constraints necessitate robust methods for identifying regions of interest in real time. This work presents an adaptive outlier detection algorithm based on the reconstruction error of Principal Component Analysis (PCA) for feature reduction, designed explicitly for space mission applications. The algorithm adapts dynamically to evolving data distributions by using Incremental PCA, enabling deployment without a predefined model for all possible conditions. A pre-scaling process normalizes each feature's magnitude while preserving relative variance within feature types. We demonstrate the algorithm’s effectiveness in detecting space plasma events, such as distinct space environments, dayside and nightside transients phenomena, and transition layers through NASA's MMS mission observations. Additionally, we apply the method to NASA's THEMIS data, successfully identifying a dayside transient using onboard-available measurements. 
\end{abstract}

\section{Introduction}
Space missions, such as NASA’s Magnetospheric Multiscale (MMS) Mission~\cite{Burch2016} and Time History of Events and Macroscale Interactions during Substorms (THEMIS)~\cite{Angelopoulos2009}, generate large volumes of multi-featured time series data from in-situ measurements. These data require efficient methods for identifying and prioritizing scientifically relevant events.
More broadly, large-scale data collection is essential across various scientific domains, including Earth-based weather monitoring~\cite{PEREZ202379} and Hyper Spectral Imaging (HSI) from Earth observation (EO) satellites~\cite{qian2021}. 
In space missions, limited onboard computational resources and restricted downlink capacity necessitate real-time event detection to prioritise the most valuable data for transmission~\cite{Burch2016}.

A challenge in analyzing space mission data is detecting outlier data points. These outliers may indicate both instrument failures and important physical phenomena occurring in the near-earth space environment, such as plasma environment crossings, magnetic reconnection events, or transient events. While some outliers may result from sensor anomalies or noise, others may reveal structures or processes not accounted for by existing models~\cite{SOUIDEN2022100463}. Traditional anomaly detection methods often assume static data distributions, which is generally not true for space plasma environments: dynamic changes in solar wind conditions and magnetospheric interactions can lead to evolving data characteristics.

One additional challenge for in-situ automatic data analysis is that onboard computational resources are largely constrained. These devices can only support lightweight and adaptive algorithms capable of processing high-dimensional data streams in real time~\cite{zamryLightweightAnomalyDetection2021}. Many existing methods require extensive prior knowledge of the data or rely on complex models that are impractical for deployment on resource-limited spacecraft. An effective approach should balance computational efficiency with the ability to detect scientifically meaningful outliers without extensive pre-training on specific datasets.

In this work, we present an outlier detection algorithm tailored for space missions. The method leverages reconstruction error from Principal Component Analysis (PCA) for feature reduction and adapts dynamically to new data distributions using Incremental PCA. This enables real-time outlier detection onboard spacecraft. Our focus is on evaluating the algorithm's principle for use in detecting outliers and demonstrating the algorithm’s effectiveness in identifying plasma events such as bow shock crossings in MMS data and foreshock bubbles in THEMIS observations. The main contributions of this work are the following:
\begin{itemize}
    \item Introduce an adaptive outlier detection algorithm based on the PCA reconstruction error, which can be applied to streaming data. The algorithm works without a pre-training step and dynamically adapts to the dominant features in new data.
    \item Extend the algorithm to handle different feature types with different magnitudes by coupling the feature scaling to a group with features of the same type. This approach preserves the relative variance within each feature group.
    \item Demonstrate the algorithm's effectiveness in finding boundary crossings and other scientifically interesting events in MMS data using multiple features.
    \item Show that the algorithm can detect scientifically relevant events in data available onboard the spacecraft using the THEMIS mission, potentially supporting real-time identification.
\end{itemize}

\section{Background}
In this work, we focus on the use-case of finding outliers in multi-featured space plasma time-series data using MMS~\cite{Burch2016}, launched in 2015 and THEMIS, launched in 2007~\cite{Angelopoulos2009}, as examples. MMS was launched to investigate magnetic reconnection in the boundary regions of Earth's magnetosphere using unprecedented time scales. It consists of four spacecraft flying in formation through the dayside magnetopause and the nightside magnetotail regions~\cite{Burch2016}. 

The THEMIS mission was launched to investigate the trigger and evolution of substorms. The mission consists of five satellites, lining up to track particles along the magnetotail. While the primary mission goal was to perform measurements in the magnetotail, in the nightside region of Earth's magnetosphere, the spacecraft also obtained measurements from the dayside region~\cite{Angelopoulos2009}. In this work, we have used an interval of dayside data from THEMIS.

Both MMS and THEMIS have limited storage capabilities, constraining the amount of data that can be collected. Furthermore, the presence of limitations in the downlink capabilities means that the collected data of the highest possible value has to be prioritized. The collection limitation and prioritization were performed using pre-specified temporal region-of-interests (ROI) with corresponding locations in space, using onboard calculated indicator values and human-made selections. These selections are intended to prioritize the most valuable mission data, especially those that are significantly different from the general state of the regions and have a high likelihood of containing interesting events~\cite{Burch2016,Angelopoulos2009}.

As spacecraft orbit Earth, they transition through multiple regions, including the magnetosphere, where Earth's magnetic field shapes the plasma environment and areas beyond its protective influence, where the solar wind and the Sun’s magnetic fields dominate. They also traverse key boundaries such as the bow shock, where the solar wind slows and deflects and the magnetopause, which separates Earth's magnetic domain from the solar wind. Classifying which region the measurements belong to can help direct scientists toward phenomena of more interest~\cite{toy-edens2024}. Determining the regions and finding the crossings between them can help to optimize the data collection onboard the spacecraft.

\subsection{Outlier Detection in Streaming Data}
Outlier detection deals with finding samples, or groups of samples, that differ from the general structure of the remaining samples. Traditionally, most outlier detection methods deal with static data where the entirety of the data range is known~\cite{yehMatrixProfile2016}. If the data is high-dimensional, feature reduction techniques, such as Principal Component Analysis (PCA)~\cite{finleyGeneralizedTimeSeriesAnalysis2024} or Autoencoders \cite{bakraniaDimReductionSpace2020}, can be utilized to simplify the problem and reduce the search space~\cite{SOUIDEN2022100463}.

However, streams of data add to the complexity of finding outliers. The temporal variability of the data means that what is considered an outlier can shift with time. The nominal values of a feature and feature importance may also shift in time~\cite{SOUIDEN2022100463}. This can render an initial data model incorrect and any outlier detection algorithm needs to be able to adapt to these shifts in the data.

\textbf{PCA}\label{sec:pca} is a linear feature reduction method that surfaces the top $N$ features, the Principal Components (PC) containing the most variance (i.e. information), from the original feature space~\cite{Ross2008}. Once in the reduced feature space, the samples can be separated into groups using different clustering techniques~\cite{bakraniaDimReductionSpace2020, angeo-39-861-2021}. Then, outliers can be located based on the distance to these clusters' centroids or the density.
Another option for finding outliers is to transform the samples back to the original feature space and evaluate how much the reconstructed features ($R_f$) differ from the original features ($F_f$). This reconstruction error ($E_f$):
\begin{equation} \label{eq:req_error}
  E_f = R_f - F_f
\end{equation}
\noindent is the loss of information from the feature reduction. Samples where features have large reconstruction errors will deviate from the PCA model and can, therefore, be considered outliers compared to the other samples~\cite{bhushanINCREMENTALPRINCIPALCOMPONENT2015a}. Incremental PCA is a variant of PCA that can be used to build the PCA model incrementally~\cite{Ross2008}. This can be necessary if the data is too large to load all the samples into memory at once, or for the use-case presented in the paper, not all the data is available when the initial model is created.

\section{Methodology}
In this work, we present a method for detecting outliers in streaming multi-feature data based on evaluating the reconstruction error stemming from feature reduction using Incremental PCA~\cite{Ross2008}. 
The reconstruction error $E_i$, for a sample $i$, is calculated as the Euclidean norm of the reconstruction errors $E_f$ for all the features $f$. 
By looking for large and rapid changes in this error, we can find samples that deviate from the model and the previous samples.

\subsection{Outlier Detection}\label{sec:algorithm}

\begin{figure}[t]
    \centering
    \includegraphics[width=.9\linewidth]{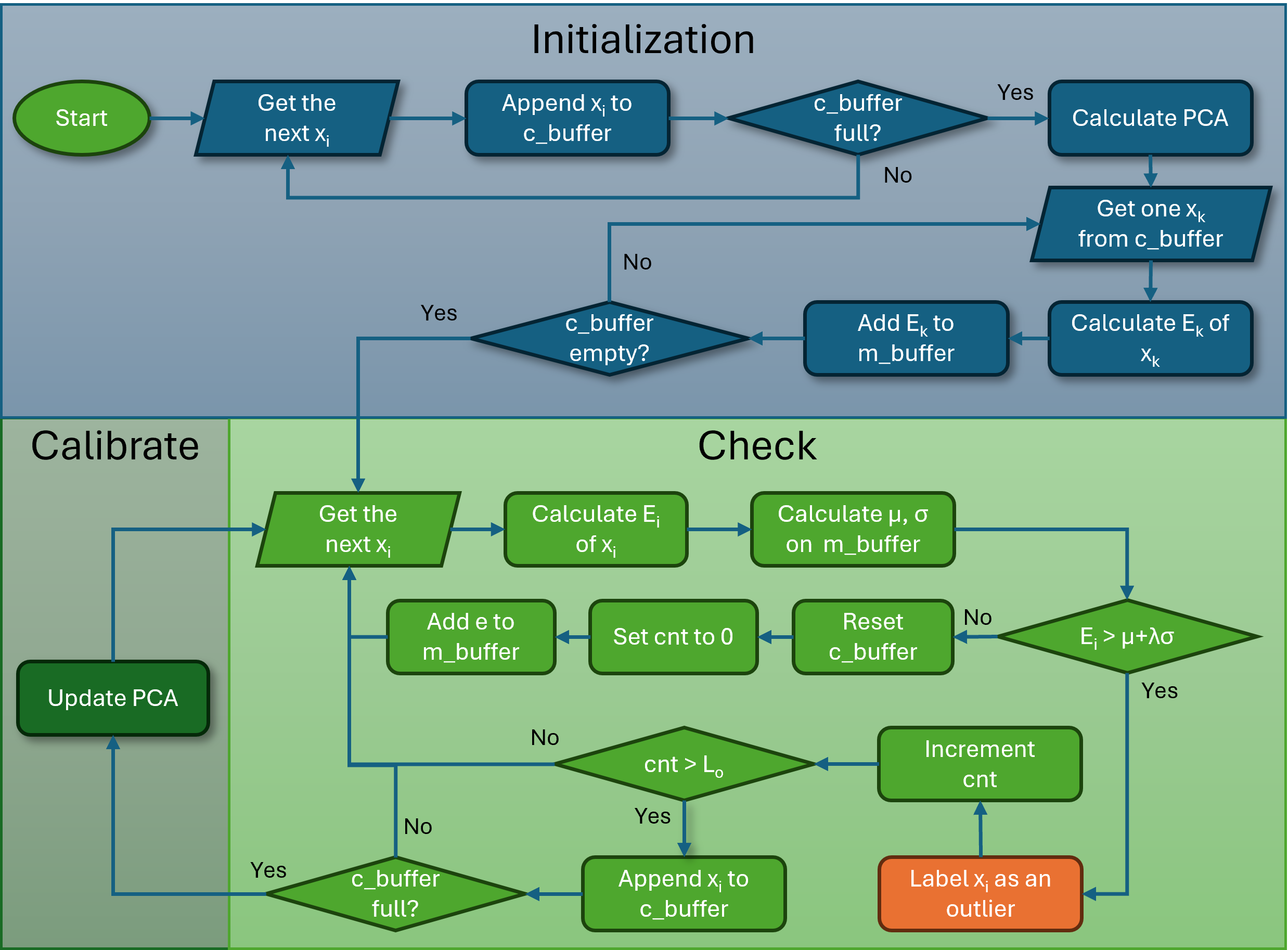}
    \caption{Flowchart describing the working principles of the adaptive algorithm.}
    \label{fig:alg_flow}
    \vspace{-1.5em}
\end{figure}

The outlier detection algorithm\footnote{\url{https://github.com/Jonah-E/multi-feature-outlier-detection}} operates in three different modes, Initialization, Check and Calibrate, as shown in Fig.~\ref{fig:alg_flow}.
In the initialization mode, (Fig.~\ref{fig:alg_flow} - top), the initial model is built based on the first samples retrieved by the algorithm.
Following the initialization mode is the check mode, which is the main operating mode (Fig.~\ref{fig:alg_flow} - bottom right). Here, samples are retrieved and evaluated for outliers. If multiple sequential samples are labeled as outliers, then a calibration using these outlier samples is initiated.
In the calibration mode, the PCA model is updated based on the latest outlier samples. 

There are five parameters controlling the execution of the algorithm: the calibration buffer size ($S_c$), the number of components ($N$) used in the PCA, the mean buffer size ($S_m$), the threshold ($\lambda$) and the outlier limit ($L_o$). $S_c$ controls the size of the calibration buffer (c\_buffer). The calibration buffer is used to store samples to use for calibration. When this buffer is full, either in the initialization mode or the check mode, the PCA is calculated using the samples in this buffer. $N$ controls how many components of the PCA are calculated and thereby the size of the reduced feature space. A lower value will, in general, mean a higher reconstruction error, but changing this parameter can also change what events are discovered by the algorithm. $S_m$ controls the size of the mean buffer (m\_buffer). The mean buffer is a fixed width circular buffer used to store the reconstruction error ($E_i$) of the latest $S_m$ samples not labeled as outliers. This buffer is then used to calculate the current mean $\mu$ and standard deviation $\sigma$.
The $\mu$ and $\sigma$ values are then used together with the threshold $\lambda$ to calculate the maximum allowed reconstruction error, or the error threshold $T_i$, according to the equation
\begin{equation}\label{eq:error_threshold}
     T_i = \mu + \lambda\sigma
\end{equation}
\noindent A sample $x_i$ with error $E_i$ exceeding $T_i$ will be labeled an outlier. The last parameter, $L_o$, controls how many samples $x_i$ can be labeled as outliers before samples are added to the calibration buffer. Setting this to a value larger than zero will exclude the $L_o$ first outlier samples from the calibration buffer. 

\textbf{Initialization mode: } is the start of the algorithm. Here, the algorithm builds the initial model by retrieving the next sample $x_i$ and then adding it to the calibration buffer. When the calibration buffer is full, the PCA is calculated using the samples in the calibration buffer. Following this, the reconstruction error ($E_k$) of the samples ($x_k$) in the calibration buffer is calculated and added to the mean buffer. Depending on the application of the algorithm, it could be beneficial to pre-calculate the PCAs on a curated dataset containing data from expected regions. Then, the initialization step could be skipped.

\textbf{Check mode: } The first step in the check mode is to retrieve the next sample $x_i$ and calculate the $E_i$ of this sample. Then $\mu$ and $\sigma$ are calculated using the samples in the m\_buffer and $T_i$ can be calculated according to Equation~\ref{eq:error_threshold}. If the $E_i$ is larger than $T_i$, then the sample is labeled as an outlier and a counter (cnt) is incremented. If this counter exceeds $L_o$, then the sample $x_i$ is also added to the calibration buffer. If the sample is not an outlier, the calibration buffer is cleared, the counter is reset to 0 and the $E_i$ is added to the mean buffer. Only samples not labeled as outliers are added to the mean buffer to prevent outlier samples from inflating the error threshold. The last step is to evaluate if the calibration buffer is full. If it is full, multiple samples in a row have been labeled as outliers and the PCA model is likely no longer correct. Then, a calibration is initiated. 

\textbf{The calibration mode: } is where the PCA is updated using the samples from the calibration buffer, bottom left in Fig.~\ref{fig:alg_flow}, before the algorithm reenters check mode and the next sample is processed.

\rowcolors{2}{gray!25}{white}
\setlength{\tabcolsep}{0.5em}
\begin{table}[ht]
    \centering
    \vspace{-1.5em}
    \caption{MMS intervals with primary ROI, dayside data intervals are originating from Ref.~\cite{Raptis2025} and nightside intervals from Ref.~\cite{richard2022}.}
    \label{tab:nightside-with-roi}
    \label{tab:dataset-with-roi}
    \begin{tabular}{c|c c c}

         Nr & Data Interval & ROI & Day or Nightside \\ \hline
         0 & 2017-12-14 16:00 $\rightarrow$ 2017-12-14 05:00 & - & Dayside\\
        1 & 2017-12-17 16:00 $\rightarrow$ 2017-12-17 22:00 & 17:52 $\rightarrow$ 17:54 & Dayside\\
        2 & 2018-01-12 00:50 $\rightarrow$ 2018-01-12 06:00 &  01:50 $\rightarrow$  01:52 & Dayside\\
        3 & 2018-12-14 02:00 $\rightarrow$ 2018-12-14 07:20 & 04:21 $\rightarrow$ 04:22 & Dayside\\

        & &  04:40  $\rightarrow$ 04:42  & \\
        \rowcolor{white}4
         &  2018-12-10 04:00 $\rightarrow$ 2018-12-10 11:00 & 05:12 $\rightarrow$ 05:25 & Dayside\\
         & & 06:27 $\rightarrow$ 06:31  & \\
        5 & 2019-01-05 16:00  $\rightarrow$ 2019-01-05 19:00 & 17:38 $\rightarrow$ 17:41& Dayside\\
        6 & 2021-01-12 00:00  $\rightarrow$ 2021-01-12 06:00 & 01:18 $\rightarrow$ 01:21 & Dayside\\
        7 & 2021-02-13 10:00  $\rightarrow$ 2021-02-13 18:00 & 11:05 $\rightarrow$ 11:06 & Dayside\\
        8 & 2022-05-02 18:00  $\rightarrow$ 2022-05-02 22:00 & 18:23 $\rightarrow$ 18:25 & Dayside\\
        9 & 2022-11-24 02:00  $\rightarrow$ 2022-11-24 09:00 & 04:16 $\rightarrow$ 04:18 & Dayside \\
        10 & 2023-01-16 06:00  $\rightarrow$ 2023-01-16 11:00 & 08:21 $\rightarrow$ 08:24 & Dayside \\
        11 & 2017-07-23 12:00 $\rightarrow$ 2017-07-23 18:00 & 16:55 $\rightarrow$ 16:56 & Nightside\\
        12 & 2021-08-14 16:00 $\rightarrow$ 2021-08-15 06:00 & 01:23 $\rightarrow$ 01:25 & Nightside\\
    \end{tabular}
    \vspace{-2em}
\end{table}

\subsection{Data and Pre-processing}
\textbf{MMS Data: }The primary data used in this paper is from MMS. Specifically, the omni-directional ion spectrum and ion velocities in GSE coordinates from the Fast Plasma Investigation (FPI) instrument~\cite{pollockFastPlasmaInvestigation2016} and the magnetic field (hereafter referred to as the B-field) from fluxgate magnetometers (FGM), part of the FIELDS instrument suite \cite{Torbert2016}, from the MMS-1 spacecraft. Table~\ref{tab:dataset-with-roi} lists eleven dayside intervals of MMS data together with some regions of interest. These intervals are either from  MMS passing into or out of the Earth's magnetosphere from the Solar wind. Transitions from the solar wind to the magnetosheath are called \textit{bow shock} crossings, while transitions from the magnetosheath to the magnetosphere are called \textit{magnetopause} crossings. The data from the FGM is collected at a significantly higher frequency, $\sim8$Hz to $\sim16$Hz, than the FPI data, $\sim0.2$Hz, we therefore down-sampled to the sample frequency of FPI. The data used is level-2 data, which is post-processed on Earth and would not be available onboard the spacecraft.

The dayside ROI in Table~\ref{tab:dataset-with-roi} are taken from \textcite{Raptis2025} and are transient phenomena upstream of the bow shock in the ion foreshock region, which are known to energize particles and cause space weather effects~\cite{kajdic2024}. Therefore, finding these, even in level-2 data, is important for further scientific research. In addition to these transient events, the data intervals also contain other phenomena, such as bow shock and magnetopause crossings, which are operationally critical to identify automatically.

We have used two nightside data intervals from MMS-1, listed in Table~\ref{tab:nightside-with-roi}, to evaluate how the algorithm generalizes to different regions. These are intervals from the nightside magnetosphere of Earth, containing two events analyzed by \textcite{richard2022}. These are fast plasma flows associated with geomagnetic disturbances and can, therefore, cause space weather effects~\cite{angelopoulos1992bursty,Sitnov2019}.

\textbf{THEMIS spacecraft} is used as an alternative source of space plasma data with different instruments compared to MMS. In particular, we use THEMIS-C probe measurements of the ion energy flux and ion velocity from the Electrostatic Analyzer (ESA), these are measurements with minimal processing, which are readily available for onboard algorithms~\cite {Angelopoulos2009,themisc-data-web}. The interval investigated is from 2008-07-15 and contains a foreshock transient~\cite{terry2019}.

\begin{figure}[t]
    \centering
    \includegraphics[width=0.8\linewidth]{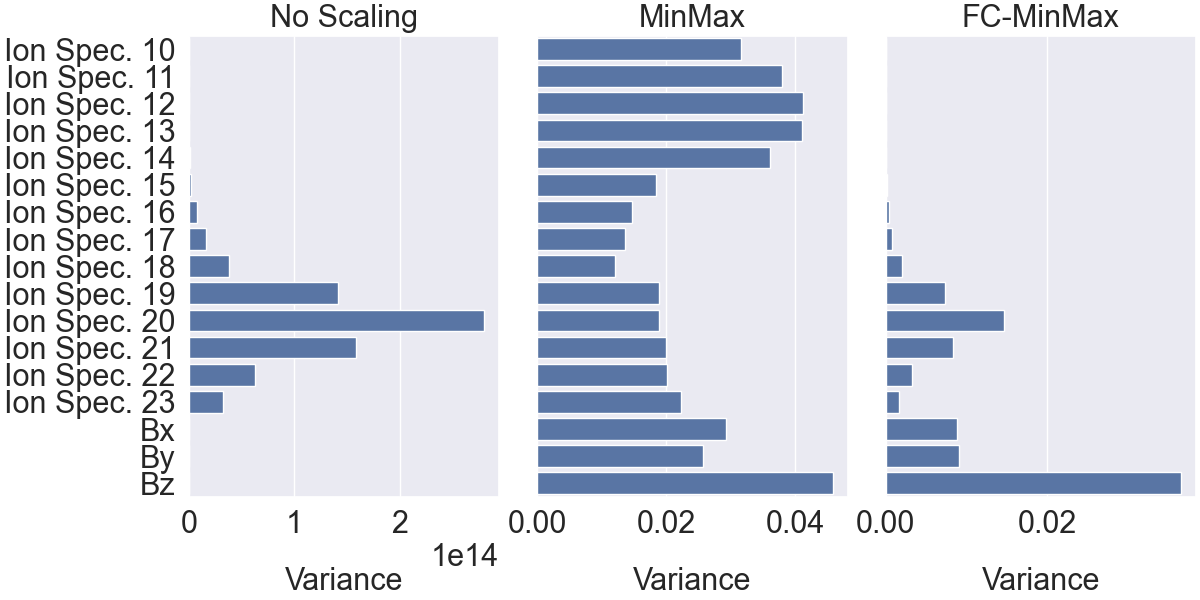}
    \caption{Comparison of the variance of the different features for no scaling (Left), MinMax scaling (Middle) and FC-MinMax scaling (Right). The features are MMS omnidirectional ion spectrum, channel 10 to 23 and B-field  from 2017-12-17, 20:00 to 21:50, while MMS is traversing the magnetosheath region.}
    \label{fig:variance_comp}
    \vspace{-1.5em}
\end{figure}

\textbf{Multiple Features:} \label{sec:scaling} The PCA method finds the axis with the highest variance; therefore, for PCA to be effective, all features must be normalized to the same range. When PCA is used in data with one type of feature, for example, the omnidirectional ion spectrum used in this paper, 
the values of all features can be expected to have the same magnitude. However, if multi-feature types are included, these new features can have values with magnitudes that differ significantly. Then, the feature types with the highest magnitudes will dominate the PCA. This can be seen in the leftmost plot of Fig.~\ref{fig:variance_comp}, where Ion spectrum channels 19 to 21 have the highest variance, while the B-field variance is not visible. A common way to solve this issue is to scale the features, for example, MinMax scaling, which scales each feature to be between zero and one. However, this destroys the relative variance within a feature type. This can be seen in the middle plot of Fig.~\ref{fig:variance_comp}, where the highest variance for the ion spectrum data has now moved to channels 12 and 13. To solve this issue, we introduce a coupling between features of the same type when scaling. This Feature Coupled MinMax (FC-MinMax) scaling scales the coupled features to the same min and max values, which are calculated over the group of features of the same type. 

The variance after using the FC-MinMax scaling can be seen in the rightmost plot of Fig.~\ref{fig:variance_comp}. Here, the relative variance within each group, ion spectrum and B-field are retained while scaling the features to have the same magnitude. We can now see that the B-field, together with the ion spectrum, will affect the PCA. In our tests, we have used data interval 0 from Table~\ref{tab:dataset-with-roi} to calculate the scaling for each feature group when testing the data from the MMS dayside intervals. For other data, we have calculated the scaling based on the specific data.

\section{Results}\label{sec:results}

\begin{figure}[t]
    \centering
    \includegraphics[width=0.9\linewidth]{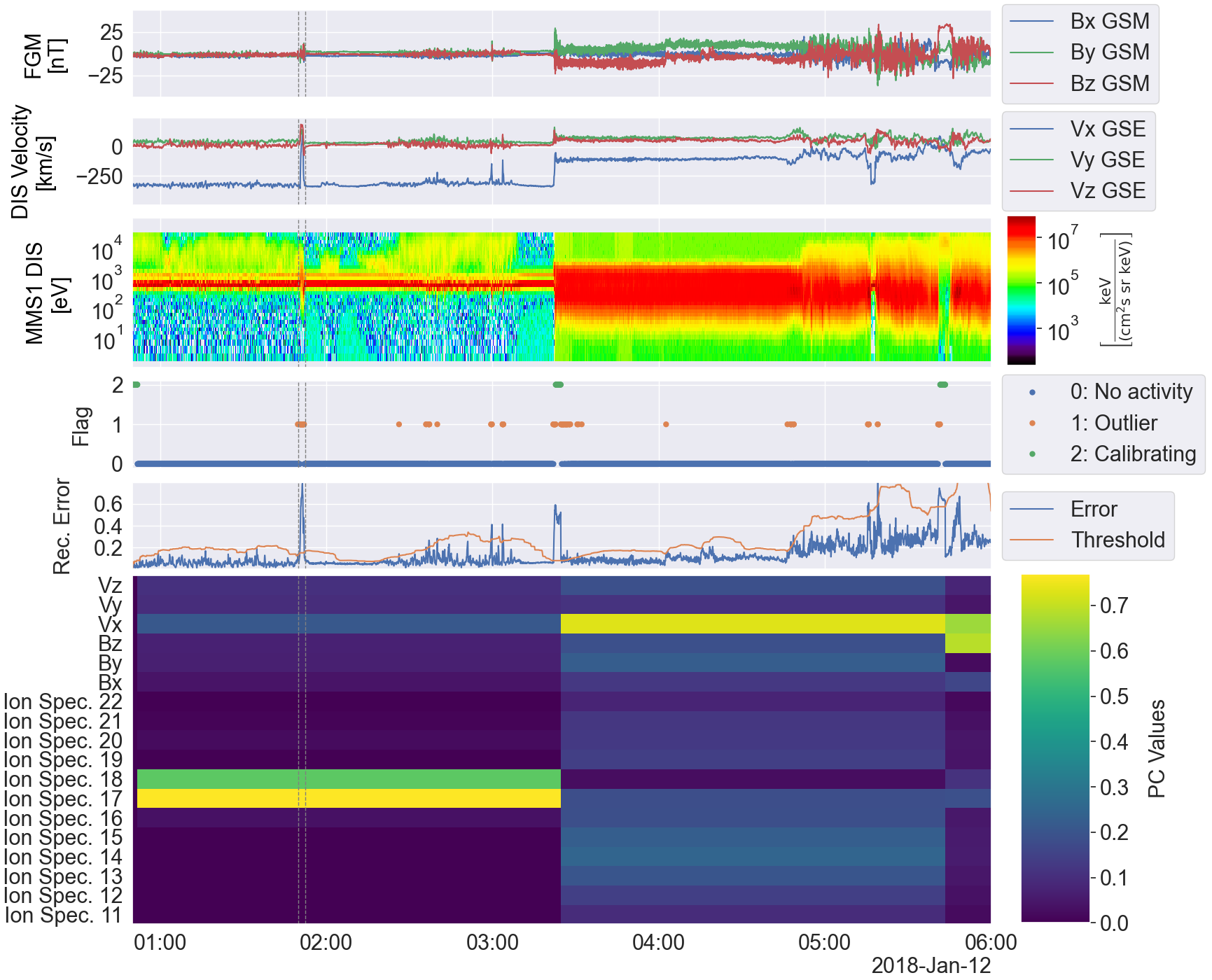}
    \caption{Algorithm applied to multi-feature data from MMS dayside interval 2, the primary ROI is marked with dashed lines, $S_c=25$, $S_m=150$, $L_o = 10$ and $\lambda=4$. The bottom plot is the principal component (PC) at each time; ion spectrum channels 0-10 and 23-31 have values close to zero and are not shown.}
    \label{fig:multi-feature-detection}
    \vspace{-1.5em}
\end{figure}

Fig.~\ref{fig:multi-feature-detection} shows the outlier detection algorithm performance when applied to a single time-series of MMS data. The features used are the ion omnidirectional spectrum, the B-field and the ion velocities. Each feature is scaled using the FC-MinMax scaling with the scaling factors calculated on data interval 0, see Table~\ref{tab:dataset-with-roi}. The bottom three plots describe the functionality of the algorithm. Samples with the flag `outlier' or `calibration' are outlier samples. However, samples with the flag `calibration' are used to update the PCA model.

The initialization stage can be seen furthest to the left in the plot; here, there are no outlier detections and the flag is set to `Calibrating'. Following this, we can see how the reconstruction error and detection threshold are updated as more samples are ingested. At 01:50, the algorithm detects ROI from Table~\ref{tab:dataset-with-roi}. The next region of interest in the data is the bow shock crossing at 03:22, where the spacecraft enters the magnetosheath. The algorithm detects the crossing by the sudden increase in the reconstruction error. However, as the error does not decrease below the threshold of 35 samples ($L_o$ + $S_c$), the entry triggers a calibration for the new region, which is shown by the samples labeled `Calibration'. After the calibration, the error decreases below the threshold. One more calibration is triggered at the short entry into the magnetosphere region at around 05:40 after a magnetopause crossing. Both of these crossings highlight different physical dynamics and their automatic finding can facilitate scientific research. 

\begin{figure}[t]
    \centering
    \includegraphics[width=0.9\linewidth]{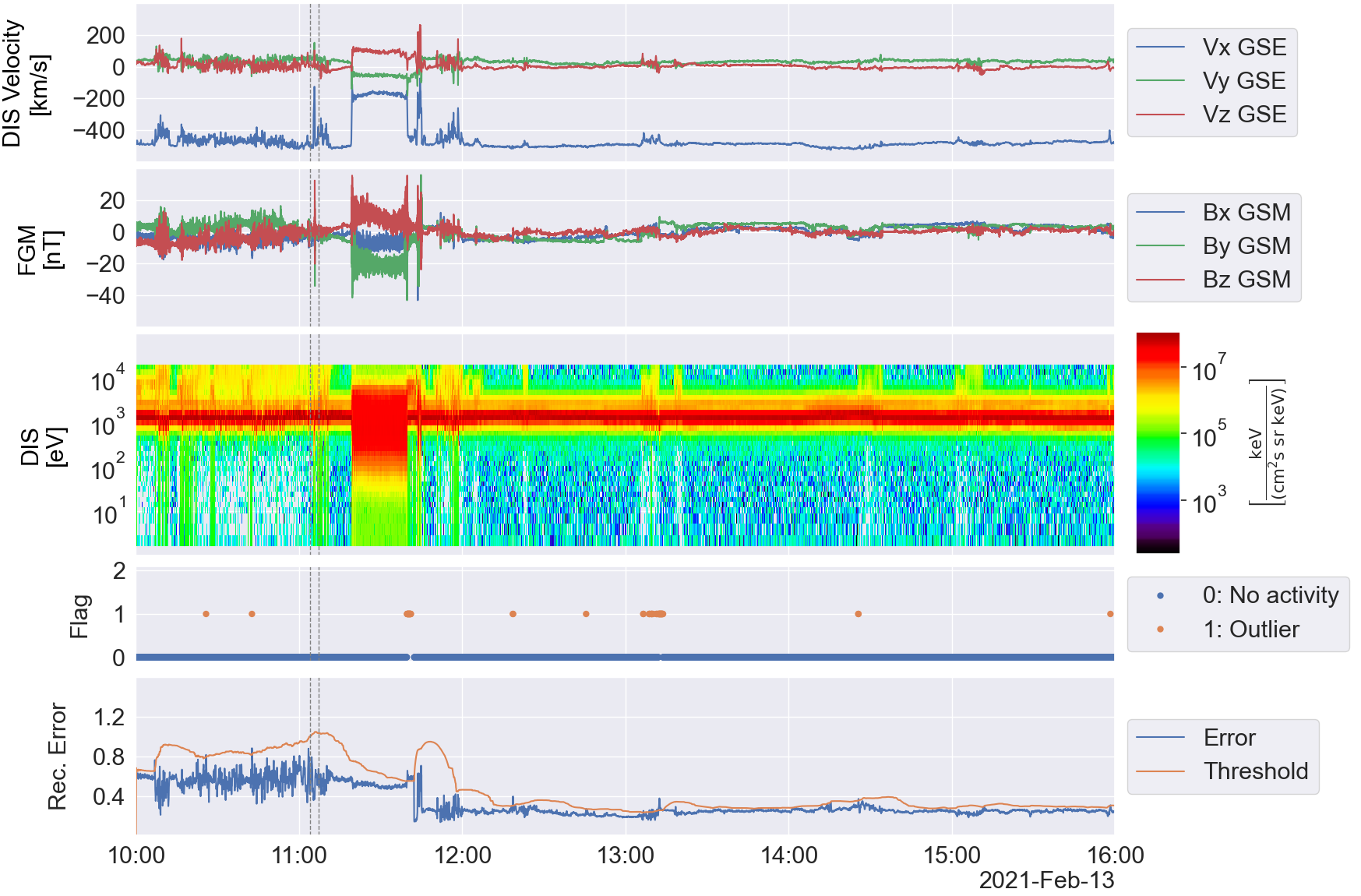}
    \caption{Example of a ROI (dashed lines) the algorithm showed no detection.}
    \label{fig:harder-data}
    \vspace{-1.5em}
\end{figure}

The bottom panel in Fig.~\ref{fig:multi-feature-detection} shows how each feature affects the principal component, inversely, this also shows which feature affects the reconstruction error the most. We can see there that in the solar wind region, the PCA model is dominated by the ion spectrum channels 17 and 18, while after entering the magnetosheath, multiple spectrum channels, together with the B-field and velocities, affect the model, where the Vx feature is now dominant. 

\textbf{Optimal parameters: }
As mentioned in \ref{sec:algorithm}, five parameters control the algorithm. We have performed a rudimentary parameter optimization on the set of dayside data intervals from Table~\ref{tab:dataset-with-roi}, maximizing the detection of the listed ROIs. For this data, with the feature types, ion spectrum, B-field and ion velocities, we have found that setting the parameters as $S_c=15$, $N=2$, $S_m=170$, $\lambda=4$ and $L_o = 20$, balances finding the ROIs with false detections. 
With these parameters, the algorithm can find nine out of the twelve dayside ROI specified in Table~\ref{tab:dataset-with-roi} when all the data intervals are fed in sequence to the algorithm without restarting it between intervals. However, the time shift between the intervals also introduces an artificial shift in the reconstruction error. When this shift is towards a larger error, the algorithm over-calibrates to the region at the start of the new interval, which can help it detect the primary ROIs. For a more realistic analysis, the mean buffer was reset to the first two samples when entering each new data interval. With these settings, most of the samples are marked as either 'No Activity' or 'Outlier', 97.8\% and 2.0\% respectively, only 0.2\% are used to update the model.

\begin{figure}[t]
    \centering
    \includegraphics[width=0.85\linewidth]{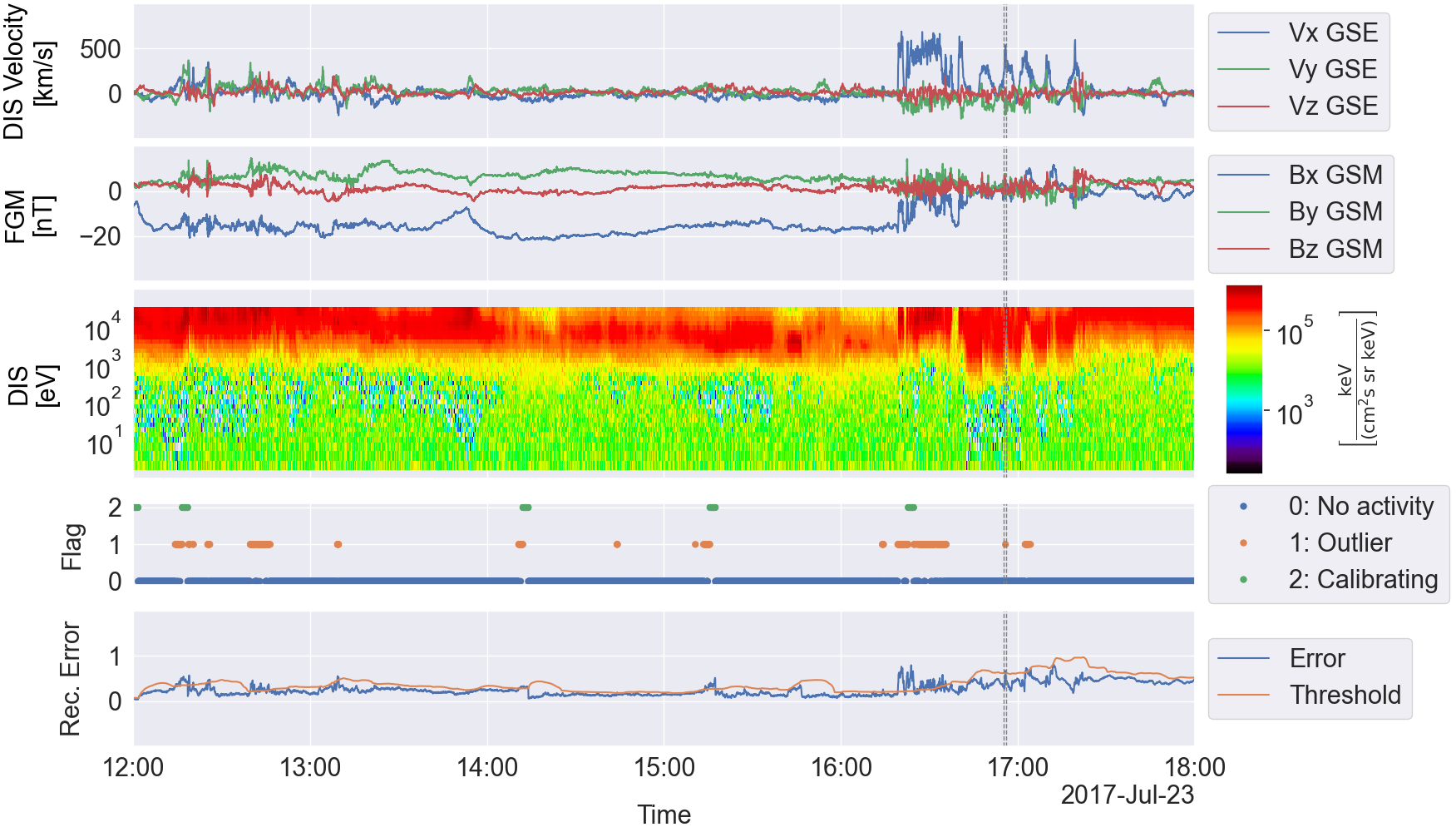}
    \vspace{-0.5em}
    \caption{The algorithm applied to MMS Nightside data interval 11 from Table~\ref{tab:nightside-with-roi}.}
    \label{fig:night-side}
    \vspace{-1.5em}
\end{figure}
The detection result for data interval 7 is plotted in Fig.~\ref{fig:harder-data}. Here, the algorithm is missing the primary ROI. However, it does find the crossing out of the magnetosheath at 11:39. The following quick crossing in and out of the magnetosheath, at 11:43 to 11:44, is, however, missed due to the sudden drop in reconstruction error from the first exit causing an increase in the standard deviation and thereby in the error threshold. 
 
\textbf{MMS-1 Nightside Data: }\label{sec:multi-comps}
When applying the algorithm to the nightside data from Table~\ref{tab:nightside-with-roi}, it can find the ROI at 2021-08-15 01:23 from the 12th data interval but not the ROI from the 11th interval. However, by increasing the number of PCAs to three and lowering the threshold to 3.5, the algorithm can find both regions. Here, the FC-MinMax scaling is calculated on the same data interval as the algorithm is applied. If scaling is instead calculated on dayside interval 0, the algorithm will over-calibrate to the ion velocities, as these have a larger value range in parts of the nightside data.
\begin{figure}[t]
    \centering
    \includegraphics[width=0.9\linewidth]{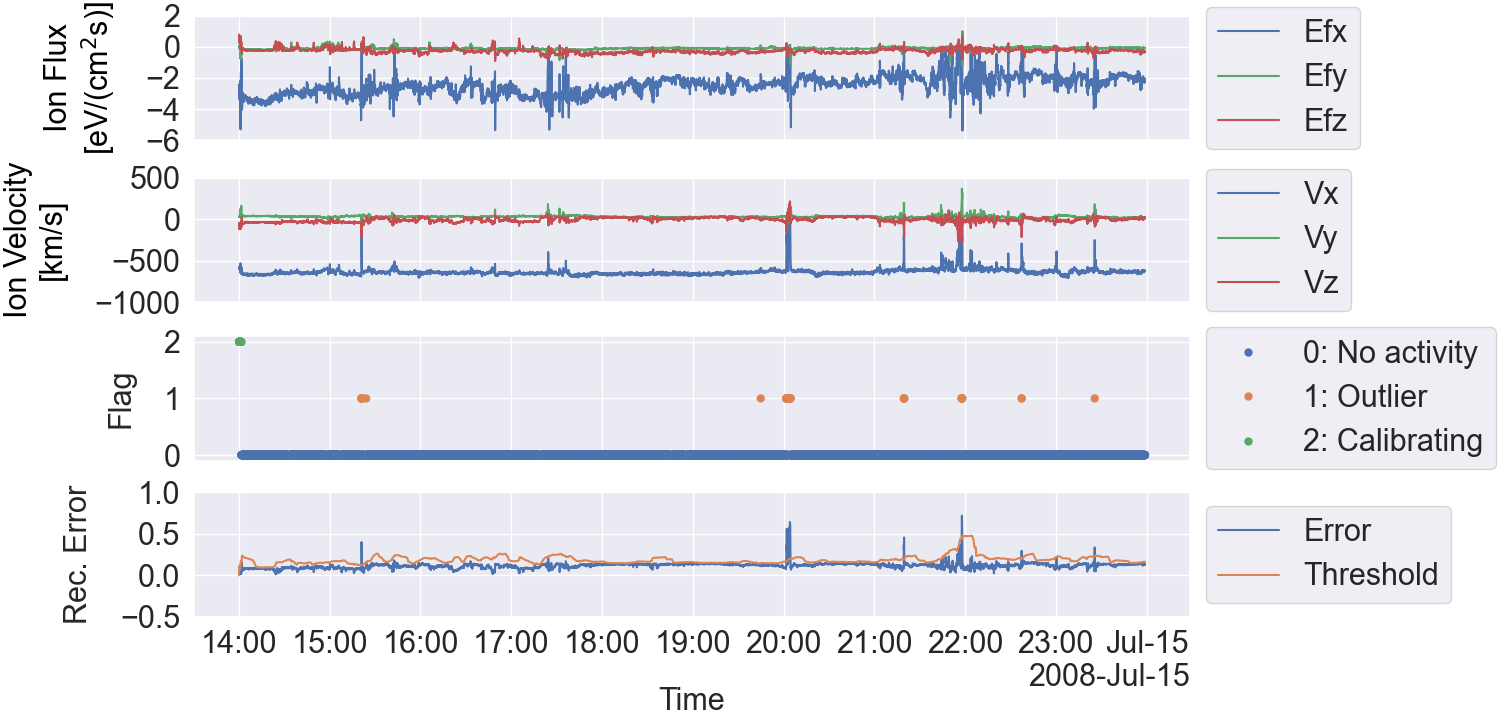}
    \vspace{-0.5em}
    \caption{The algorithm applied to an interval of dayside data from THEMIS C.}
    \label{fig:themis}
    \vspace{-1.5em}
\end{figure}

\textbf{THEMIS C Dayside Data: }
Applying the algorithm to THEMIS C intervals, with a scaling calculated on the same interval, we find several possible interesting events; The longest of these is between 20:01:51 to 20:04:36 and 21:57:44 to 21:58:08. This second event starting at 21:57:44 has been noted in previous works as a foreshock bubble~\cite{wilson2016,terry2019}.
The start and stop times indicated by our algorithm also correspond well to the start of expansion to end of compression, shown by \textcite{terry2019} in Fig.~2. For this data, the calibration batch and mean window size have been increased to 30 and 180 due to the higher sampling frequency of the THEMIS C data. 

For the full day, 2008-07-15, the first twelve hours of the data have values with significantly smaller magnitudes than the region in Fig.~\ref{fig:themis}. When applying the algorithm to this wider data interval, with a scaling calculated on the full interval, the algorithm detects the start of the new region but does not trigger a recalibration. Lowering the threshold to 3.5 causes the algorithm to more aggressively recalibrate during the transition and find the events in Fig.~\ref{fig:themis}

\section{Related Work}

\textbf{Plasma region and Bow Shock identification: } In the area of space plasma physics, bow shock crossings, which are coupled with the transition between the solar wind region and the magnetosheath are areas of active research. There have been numerous works developing methods for identifying plasma regions (e.g., solar wind and magnetosheath) and their transitions (e.g., bow shock). Both \textcite{olshevsky2021} and \textcite{breuillardAutomaticClassificationPlasma2020} utilized neural-network-based approaches to identify different plasma regions. The bow shock transitions are either identified by where network output indicates a high probability for neighboring regions~\cite{olshevsky2021} or as specific labels~\cite{breuillardAutomaticClassificationPlasma2020}. 
These approaches are powerful and displayed high accuracies on the test sets. However, as the methods rely on having a training dataset, they cannot adapt to changing input data. While the algorithm presented in this paper can benefit from samples of relevant regions to create an initial model, the algorithm does not require a training set. Instead, it can be applied directly and will adapt to new data. 

More recent work by \textcite{toy-edensClassifying8Years2024} utilized a Gaussian Mixture Model to classify the MMS dayside region, which displayed high accuracy using a method more lightweight than a full neural network.
\textcite{bakraniaDimReductionSpace2020} used a pipeline of autoencoder, PCA and Agglomerative Clustering to classify data from the ESA's Cluster spacecraft \cite{escoubet2001introduction}. \textcite{angeo-39-861-2021} utilizes PCA, self-organizing maps and K-means clustering to process multiple different quantities, such as B-field, particle velocity and particle density, to obtain a classification of the plasma region.

\textbf{PCA-based outlier detection} The two-step approach, presented by \textcite{finleyGeneralizedTimeSeriesAnalysis2024}, analyzes a region of MMS data for interesting events by subdividing the region into smaller windows and performing a PCA on the resulting matrix of subregions. Then, a One-Classifier support vector machine (OCSVN) is used to find the windows containing outliers. This technique shows promising results; however, it is very computationally expensive, as the PCA and OCSVN have to be recomputed for each analyzed region. \textcite{zamryLightweightAnomalyDetection2021} utilizes a PCA feature reduction coupled with a One-Class Support Vector Machine to detect anomalous sensor readings in data from a Wireless Sensor Network. \textcite{bhushanINCREMENTALPRINCIPALCOMPONENT2015a} applied an Incremental PCA on streaming data with geospatial spread. They only consider finding outliers in one type of feature from one sensor type and the outliers are artificially introduced into the data. 
Furthermore, the PCA model is constantly updated, increasing the computational complexity. The algorithm presented in this paper limits the computational complexity by only updating the PCA model when necessary. 

\section{Discussion and Conclusion}
In this work, we have presented a lightweight outlier detection algorithm for streaming multi-feature data. The algorithm can adapt to new data regions, enabling its deployment in applications where all the data regions are not known or fully understood. We utilize the Incremental PCA to have a lightweight algorithm that could be deployed on low-performing hardware. However, the basic principles in the algorithm can be applied using other unsupervised feature reduction techniques, where a reconstruction error can be calculated, for example, with autoencoders. Furthermore, while we have demonstrated the algorithm using space mission data, it can also be used to analyze other types of streaming data, such as measurements from a weather monitoring station. This will have to be evaluated for the specific data in question. 

With optimal parameters, the algorithm can find most dayside ROIs, together with several crossings from solar wind and ion foreshock to the magnetosheath. 
For the nightside data, the number of components needed to be increased and the threshold slightly lowered. If we compare where our primary ROI is located in the dayside data to where it is located in the nightside data, we can understand why. In the dayside data, the ROIs are located in the Solar Wind/Ion foreshock regions, where the ion spectrum mainly consists of a tight beam of energy, as we can see in Fig.~\ref{fig:multi-feature-detection}. This region is well-defined by only two parameters and outliers are visible in the reconstruction error. The nightside data is more varied around the marked ROI, leading to the ROI being masked by other errors. By increasing the number of components, the model can better describe the structure of the data, allowing the ROI to be found. This result indicates that the algorithm can also be used as a data mining tool to find interesting events in the dayside and nightside regions of the Earth's magnetosphere. This can help scientists find events for further studies. However, the different parts of the nightside can vary significantly and further studies will be necessary to understand how well the algorithm performs on data from the nightside region. 

The algorithm can also be used on data from different spacecraft, as demonstrated by the use of THEMIS C data. In this case, the data used would be available onboard the spacecraft, indicating the algorithm's potential to be used onboard to inform the decision on which data to prioritize for downlink. In a real mission scenario, the algorithm would be applied to data which has already been subjected to a robust onboard pre-processing check for sensor failures. However, the algorithm needs to be more extensively tested using data available onboard, from multiple spacecraft, to show that the principle is robust enough for use in future space missions. Furthermore, the algorithm has to be evaluated on representative hardware before it can be used onboard during a space mission. Therefore, future work will focus on the creation and evaluation of a version of the algorithm optimized for low-performing hardware.

The FC-MinMax scaling used to scale the different feature types enables the use of multiple feature types with different magnitudes. It also opens up the possibility of adding a feature selection by weighing the different feature types differently. 
However, the necessity of this pre-scaling step is one of the two main weaknesses of the current algorithm, the second is how the threshold is defined. To further advance the algorithm, we will focus on this pre-scaling step and evaluate ways to integrate it into the algorithm itself. It would then be calibrated at the same time as the algorithm in the initialization stage. During a recalibration, the new data can be checked against the existing scaler to see if it needs to be updated. However, this could mean that the PCA model has to be recalculated and not just updated using the new samples.

For the second weakness, the main issue is that defining the threshold using the standard deviation can lead to large thresholds when there are large decreases in the reconstruction error, as seen in Fig.~\ref{fig:harder-data}. This can lead to missed detections of interesting events. Furthermore, an incorrectly selected threshold can lead to excessive recalibration of the algorithm. To address these issues, the threshold could be dynamically updated either continuously or during calibration. Another option to limit excessive recalibration could be to introduce a forgetting factor to the IncrementalPCA. This would decrease the importance of earlier data and allow the new data to be more prominent in the model~\cite{Ross2008}.

The algorithm presented in this paper is a promising approach to finding outliers in multi-feature data. It highlights areas of interest for further investigation and can inform decisions on which data should be prioritized for downlink. To further prioritize the data, the algorithm's output can be evaluated for the number of sequential outliers and the height of the error signal above the threshold.

\printbibliography

@article{breuillardAutomaticClassificationPlasma2020,
  title = {Automatic {{Classification}} of {{Plasma Regions}} in {{Near-Earth Space With Supervised Machine Learning}}: {{Application}} to {{Magnetospheric Multi Scale}} 2016--2019 {{Observations}}},
  shorttitle = {Automatic {{Classification}} of {{Plasma Regions}} in {{Near-Earth Space With Supervised Machine Learning}}},
  author = {Breuillard, Hugo and Dupuis, Romain and Retino, Alessandro and Le Contel, Olivier and Amaya, Jorge and Lapenta, Giovanni},
  year = {2020},
  journal = {Frontiers in Astronomy and Space Sciences},
  volume = {7},
}

@inproceedings{escoubet2001introduction,
  title={Introduction the cluster mission},
  author={Escoubet, CP and Fehringer, Michael and Goldstein, Melvyn},
  booktitle={Annales Geophysicae},
  volume={19},
  year={2001},
  organization={Copernicus GmbH}
}

@article{angelopoulos1992bursty,
  title={Bursty bulk flows in the inner central plasma sheet},
  author={Angelopoulos, Vassilis and Baumjohann, Wo and Kennel, CF and Coroniti, F Vo and Kivelson, MG and Pellat, R and Walker, RJ and L{\"u}hr, H and Paschmann, G},
  journal={Journal of Geophysical Research: Space Physics},
  volume={97},
  number={A4},
  pages={4027--4039},
  year={1992},
  publisher={Wiley Online Library}
}

@article{finleyGeneralizedTimeSeriesAnalysis2024,
  title = {Generalized {{Time-Series Analysis}} for {{In Situ Spacecraft Observations}}: {{Anomaly Detection}} and {{Data Prioritization Using Principal Components Analysis}} and {{Unsupervised Clustering}}},
  shorttitle = {Generalized {{Time-Series Analysis}} for {{In Situ Spacecraft Observations}}},
  author = {Finley, Matthew G. and {Martinez-Ledesma}, Miguel and Paterson, William R. and Argall, Matthew R. and Miles, David M. and Dorelli, John C. and Zesta, Eftyhia},
  year = {2024},
  journal = {Earth and Space Science},
  volume = {11},
}

@article{toy-edensClassifying8Years2024,
  title = {Classifying 8 {{Years}} of {{MMS Dayside Plasma Regions}} via {{Unsupervised Machine Learning}}},
  author = {{Toy-Edens}, Vicki and Mo, Wenli and Raptis, Savvas and Turner, Drew L.},
  year = {2024},
  journal = {Journal of Geophysical Research: Space Physics},
  volume = {129},
}

@article{zamryLightweightAnomalyDetection2021,
  title = {Lightweight {{Anomaly Detection Scheme Using Incremental Principal Component Analysis}} and {{Support Vector Machine}}},
  author = {Zamry, Nurfazrina M. and Zainal, Anazida and Rassam, Murad A. and Alkhammash, Eman H. and Ghaleb, Fuad A. and Saeed, Faisal},
  year = {2021},
  journal = {Sensors},
  volume = {21},
}

@article{Ross2008,
author={Ross, David A.
and Lim, Jongwoo
and Lin, Ruei-Sung
and Yang, Ming-Hsuan},
title={Incremental Learning for Robust Visual Tracking},
journal={International Journal of Computer Vision},
year={2008},
volume={77},
}

@article{olshevsky2021,
author = {Olshevsky, Vyacheslav and Khotyaintsev, Yuri V. and Lalti, Ahmad and Divin, Andrey and Delzanno, Gian Luca and Anderzén, Sven and Herman, Pawel and Chien, Steven W. D. and Avanov, Levon and Dimmock, Andrew P. and Markidis, Stefano},
title = {Automated Classification of Plasma Regions Using 3D Particle Energy Distributions},
journal = {Journal of Geophysical Research: Space Physics},
volume = {126},
year = {2021}
}

@Article{angeo-39-861-2021,
    AUTHOR = {Innocenti, M. E. and Amaya, J. and Raeder, J. and Dupuis, R. and Ferdousi, B. and Lapenta, G.},
    TITLE = {Unsupervised classification of simulated magnetospheric regions},
    JOURNAL = {Annales Geophysicae},
    VOLUME = {39},
    YEAR = {2021},
}

@ARTICLE{bakraniaDimReductionSpace2020,
    AUTHOR={Bakrania, Mayur R.  and Rae, I. Jonathan  and Walsh, Andrew P.  and Verscharen, Daniel  and Smith, Andy W. },
    TITLE={Using Dimensionality Reduction and Clustering Techniques to Classify Space Plasma Regimes},
    JOURNAL={Frontiers in Astronomy and Space Sciences},
    VOLUME={7},
    YEAR={2020},
}

@Article{Burch2016,
    author={Burch, J. L.
    and Moore, T. E.
    and Torbert, R. B.
    and Giles, B. L.},
    title={Magnetospheric Multiscale Overview and Science Objectives},
    journal={Space Science Reviews},
    year={2016},
    volume={199},
}

@incollection{PEREZ202379,
title = {Chapter 5 - Oceanographic buoys: Providing ocean data to assess the accuracy of variables derived from satellite measurements},
editor = {Nicholas R. Nalli},
booktitle = {Field Measurements for Passive Environmental Remote Sensing},
publisher = {Elsevier},
year = {2023},
author = {Renellys C. Perez and Gregory R. Foltz and Rick Lumpkin and Jianwei Wei and Kenneth J. Voss and Michael Ondrusek and Menghua Wang and Mark A. Bourassa},
}

@ARTICLE{qian2021,
  author={Qian, Shen-En},
  journal={IEEE Journal of Selected Topics in Applied Earth Observations and Remote Sensing},
  title={Hyperspectral Satellites, Evolution, and Development History},
  year={2021},
  volume={14},
}

@article{pollockFastPlasmaInvestigation2016,
  title = {Fast {{Plasma Investigation}} for {{Magnetospheric Multiscale}}},
  author = {Pollock, C. and Moore, T. and Jacques, A. and Burch, J. and Gliese, U. and Saito, Y. and Omoto, T. and Avanov, L. and Barrie, A. and Coffey, V. and Dorelli, J. and Gershman, D. and Giles, B. and Rosnack, T. and Salo, C. and Yokota, S. and Adrian, M. and Aoustin, C. and Auletti, C. and Aung, S. and Bigio, V. and Cao, N. and Chandler, M. and Chornay, D. and Christian, K. and Clark, G. and Collinson, G. and Corris, T. and De~Los~Santos, A. and Devlin, R. and Diaz, T. and Dickerson, T. and Dickson, C. and Diekmann, A. and Diggs, F. and Duncan, C. and {Figueroa-Vinas}, A. and Firman, C. and Freeman, M. and Galassi, N. and Garcia, K. and Goodhart, G. and Guererro, D. and Hageman, J. and Hanley, J. and Hemminger, E. and Holland, M. and Hutchins, M. and James, T. and Jones, W. and Kreisler, S. and Kujawski, J. and Lavu, V. and Lobell, J. and LeCompte, E. and Lukemire, A. and MacDonald, E. and Mariano, A. and Mukai, T. and Narayanan, K. and Nguyan, Q. and Onizuka, M. and Paterson, W. and Persyn, S. and Piepgrass, B. and Cheney, F. and Rager, A. and Raghuram, T. and Ramil, A. and Reichenthal, L. and Rodriguez, H. and Rouzaud, J. and Rucker, A. and Saito, Y. and Samara, M. and Sauvaud, J.-A. and Schuster, D. and Shappirio, M. and Shelton, K. and Sher, D. and Smith, D. and Smith, K. and Smith, S. and Steinfeld, D. and Szymkiewicz, R. and Tanimoto, K. and Taylor, J. and Tucker, C. and Tull, K. and Uhl, A. and Vloet, J. and Walpole, P. and Weidner, S. and White, D. and Winkert, G. and Yeh, P.-S. and Zeuch, M.},
  year = {2016},
  journal = {Space Science Reviews},
  volume = {199},
}

@Article{Torbert2016,
author={Torbert, R. B.
and Russell, C. T.
and Magnes, W.
and Ergun, R. E.
and Lindqvist, P.-A.
and LeContel, O.
and Vaith, H.
and Macri, J.
and Myers, S.
and Rau, D.
and Needell, J.
and King, B.
and Granoff, M.
and Chutter, M.
and Dors, I.
and Olsson, G.
and Khotyaintsev, Y. V.
and Eriksson, A.
and Kletzing, C. A.
and Bounds, S.
and Anderson, B.
and Baumjohann, W.
and Steller, M.
and Bromund, K.
and Le, Guan
and Nakamura, R.
and Strangeway, R. J.
and Leinweber, H. K.
and Tucker, S.
and Westfall, J.
and Fischer, D.
and Plaschke, F.
and Porter, J.
and Lappalainen, K.},
title={The FIELDS Instrument Suite on MMS: Scientific Objectives, Measurements, and Data Products},
journal={Space Science Reviews},
year={2016},
volume={199},
}

@inproceedings{bhushanINCREMENTALPRINCIPALCOMPONENT2015a,
  title = {Incremental principal component analysis based outlier detection methods for spatiotemporal data streams},
  booktitle = {{{ISPRS Annals}} of the {{Photogrammetry}}, {{Remote Sensing}} and {{Spatial Information Sciences}}},
  author = {Bhushan, A. and Sharker, M. H. and Karimi, H. A.},
  year = {2015},
  publisher = {University of Pittsburgh},
}

@article{richard2022,
author = {Richard, L. and Khotyaintsev, Yu. V. and Graham, D. B. and Russell, C. T.},
title = {Are Dipolarization Fronts a Typical Feature of Magnetotail Plasma Jets Fronts?},
journal = {Geophysical Research Letters},
volume = {49},
year = {2022}
}

@article{terry2019,
author = {Terry Z. Liu  and Vassilis Angelopoulos  and San Lu },
title = {Relativistic electrons generated at Earth’s quasi-parallel bow shock},
journal = {Science Advances},
volume = {5},
year = {2019},
}

@article{wilson2016,
  title = {Relativistic Electrons Produced by Foreshock Disturbances Observed Upstream of Earth's Bow Shock},
  author = {Wilson, L. B. and Sibeck, D. G. and Turner, D. L. and Osmane, A. and Caprioli, D. and Angelopoulos, V.},
  journal = {Phys. Rev. Lett.},
  volume = {117},
  year = {2016},
}

@Inbook{Angelopoulos2009,
author="Angelopoulos, V.",
editor="Burch, J. L.
and Angelopoulos, V.",
title="The THEMIS Mission",
bookTitle="The THEMIS Mission",
year="2009",
publisher="Springer New York",
}

@misc{themisc-data-web,
    key = {Themis},
    title = {Variable Descriptions},
    url={https://themis.igpp.ucla.edu/var_desc.shtml},
    note = {Accessed: 2025-02-23}
}

@article{SOUIDEN2022100463,
title = {A survey of outlier detection in high dimensional data streams},
journal = {Computer Science Review},
volume = {44},
year = {2022},
author = {Imen Souiden and Mohamed Nazih Omri and Zaki Brahmi},
}

@INPROCEEDINGS{yehMatrixProfile2016,
  author={Yeh, Chin-Chia Michael and Zhu, Yan and Ulanova, Liudmila and Begum, Nurjahan and Ding, Yifei and Dau, Hoang Anh and Silva, Diego Furtado and Mueen, Abdullah and Keogh, Eamonn},
  booktitle={2016 IEEE 16th International Conference on Data Mining (ICDM)},
  title={Matrix Profile I: All Pairs Similarity Joins for Time Series: A Unifying View That Includes Motifs, Discords and Shapelets},
  year={2016},
}

@Article{Raptis2025,
author={Raptis, Savvas
and Lalti, Ahmad
and Lindberg, Martin
and Turner, Drew L.
and Caprioli, Damiano
and Burch, James L.},
title={Revealing an unexpectedly low electron injection threshold via reinforced shock acceleration},
journal={Nature Communications},
year={2025},
volume={16},
}

@ARTICLE{kajdic2024,
AUTHOR={Kajdič, Primož  and Blanco-Cano, Xóchitl  and Turc, Lucile  and Archer, Martin  and Raptis, Savvas  and Liu, Terry Z.  and Pfau-Kempf, Yann  and LaMoury, Adrian T.  and Hao, Yufei  and Escoubet, Philippe C.  and Omidi, Nojan  and Sibeck, David G.  and Wang, Boyi  and Zhang, Hui  and Lin, Yu },
TITLE={Transient upstream mesoscale structures: drivers of solar-quiet space weather},
JOURNAL={Frontiers in Astronomy and Space Sciences},
VOLUME={11},
YEAR={2024},
}

@article{toy-edens2024,
author = {Toy-Edens, Vicki and Mo, Wenli and Raptis, Savvas and Turner, Drew L.},
title = {Classifying 8 Years of MMS Dayside Plasma Regions via Unsupervised Machine Learning},
journal = {Journal of Geophysical Research: Space Physics},
volume = {129},
year = {2024}
}

@Article{Sitnov2019,
author={Sitnov, Mikhail
and Birn, Joachim
and Ferdousi, Banafsheh
and Gordeev, Evgeny
and Khotyaintsev, Yuri
and Merkin, Viacheslav
and Motoba, Tetsuo
and Otto, Antonius
and Panov, Evgeny
and Pritchett, Philip
and Pucci, Fulvia
and Raeder, Joachim
and Runov, Andrei
and Sergeev, Victor
and Velli, Marco
and Zhou, Xuzhi},
title={Explosive Magnetotail Activity},
journal={Space Science Reviews},
year={2019},
volume={215},
}

\end{document}